\def\year{2021}\relax
\documentclass[letterpaper]{article} 

\usepackage{booktabs} 
\usepackage{algorithm}
\usepackage{algorithmic}
\usepackage{multirow}
\usepackage{wrapfig,lipsum,booktabs}
\usepackage{subcaption}

\usepackage{aaai21}  
\usepackage{times}  
\usepackage{helvet} 
\usepackage{courier}  
\usepackage[hyphens]{url}  
\usepackage{graphicx} 
\usepackage{natbib}  
\usepackage{caption} 
\title{IEO: Intelligent Evolutionary Optimisation \\for Hyperparameter Tuning}

\author {
        Yuxi Huan,\textsuperscript{\rm 1}
        Fan Wu, \textsuperscript{\rm 1}
        Michail Basios, \textsuperscript{\rm 1}\\
        Leslie Kanthan, \textsuperscript{\rm 1}
        Lingbo Li, \textsuperscript{\rm 1}
        Baowen Xu \textsuperscript{\rm 2} \\
}
\affiliations {
    \textsuperscript{\rm 1} Turing Intelligence Technology, UK \\
    \textsuperscript{\rm 2} Nanjing University, China \\
    {yuxi,fan,mike,leslie,lingbo}@turintech.ai, bwxu@nju.edu.cn
}

\begin{document}
\maketitle

\begin{abstract}
Hyperparameter optimisation is a crucial process in searching the optimal machine learning model. 
The efficiency of finding the optimal hyperparameter settings has been a big concern in recent researches since the optimisation process could be time-consuming, especially when the objective functions are highly expensive to evaluate.
In this paper, we introduce an intelligent evolutionary optimisation algorithm which applies machine learning technique to the traditional evolutionary algorithm to accelerate the overall optimisation process of tuning machine learning models in classification problems.
We demonstrate our Intelligent Evolutionary Optimisation (\emph{IEO})  in a series of controlled experiments, comparing with traditional evolutionary optimisation in hyperparameter tuning. 
The empirical study shows that our approach accelerates the optimisation speed by $30.40\%$ on average and up to $77.06\%$ in the best scenarios.
\end{abstract}

%
%




\section{Introduction}

Hyperparameter tuning plays an important role in the process of training an optimal machine learning model. During the training process, the performance of the target model is evaluated by monitoring metrics such as the values of the loss function or the accuracy score on the test/validation set,  on which basis the hyperparameters can be fine-tuned to improve the model efficiency.

However, it usually takes a significant amount of time to find the optimal set of hyperparameters, especially for the case where a model has a large number of hyperparameters (hundreds of hyperparameters can be tuned in some models), or where the fitness function is expensive to execute, causing optimisation methods such as evolutionary algorithms, Bayesian optimisation, grid search, random search etc. to hardly scale in hyperparameter tuning. 
In other words, each time the hyperparameters are adjusted, the target model has to be retrained to evaluate the model performance, which is very inefficient when the model has a high level of complexity and expenses~\cite{coates2011analysis}.

Many recent researches have been devoting on finding better algorithms to shorten the optimisation process of hyperparameter tuning. Random search is shown effective at finding good regions for sensitive hyperparameters~\cite{jaderberg2017population}. 
Sequential Model based optimisation (SMBO)~\cite{10.1007/978-3-642-25566-3_40} has been proven improved ability to deal with randomness and reduce the computational overhead~\cite{NIPS2012_4522}.
In addition, reinforcement learning is also regarded as an effective approach for finding the optimal hyperparameter set, especially for architecture network design~\cite{jomaa2019hyprl}. 
Other than that, evolutionary algorithm has been widely used as a multi-objective approach to optimise the hyperparameter set in machine learning models and has been proven having an efficient and outperforming performance~\cite{4378745}.

In this paper, we propose an intelligent evolutionary algorithm optimisation which combines the logic of traditional evolutionary algorithm with machine learning technique to intelligently understand the pattern of under-performing hyperparameters. 
This enables the ability of skipping unnecessary model training, which results in saving the execution time while maintaining the comparable performance in terms of convergence in classification problems. 

The primary contributions of this article are summarised as follows:
\begin{itemize}
    \item 
    Improve the optimisation efficiency. 
    We incorporate our approach with evolutionary algorithm and compared the execution time of the original optimisation process with the proposed approach Intelligent Evolutionary Optimisation (\emph{IEO}) in the case of executing the same iterations.
    The results show that the proposed approach can accelerate the optimisation speed up by $4.36$ times.
    \item
    The proposed approach can keep the comparable performance with the original optimisation process.
    The experimental results show that there is no significant difference between the optimal solutions generated by original evolutionary algorithm and \emph{IEO}. 
    Therefore, while our algorithm improves efficiency, it will not affect the original optimisation algorithm’s performance.
    \item
    The proposed algorithm shares the same interface as original Evolutionary Algorithm.
    The primary logic of evolutionary algorithm is not changed.
    Therefore, users can improve the proposed algorithm easily without many changes.
\end{itemize}

\section{Related Work}
There are many different techniques that have been invented to solve the hyperparameter tuning problem~\cite{NIPS2011_4443}.
Two of the most widely used hyperparameter tuning techniques are grid search~\cite{pmlr-v97-ndiaye19a} and random search~\cite{JMLR:v13:bergstra12a}.
In grid search, a.k.a brute force search, a grid of hyperparamters are set up for evaluation, which covers every combination of hyperparameters.
The disadvantage of this approach is that the grid grows exponentially with the number of hyperparameters. 
Unlike grid search, random search select random combinations of hyperparmaters to evaluate, which over the same domain is able to find models that are as good or better within a small fraction of the computation time. 
It finds better models by effectively searching a larger, high dimensional space, less promising configuration space~\cite{JMLR:v13:bergstra12a} and also has been shown to be sufficiently efficient for learning neural networks for several datasets~\cite{NIPS2011_4443}. 

In order to improve the efficiency of hyperparameter tuning process, Sequential Model based optimisation (SMBO)~\cite{10.1007/978-3-642-25566-3_40} has been used in many applications where evaluation of the fitness function is expensive. 
The most typical and widely used one is Bayesian optimisation, which is assuming that the unknown function was sampled from a Gaussian process and maintains a posterior distribution for this function as observations are made~\cite{NIPS2012_4522}. 
In terms of picking the hyperparameters of the next experiment, one can optimise the expected improvement (EI)~\cite{srinivas2009gaussian} over the current best result or the Gaussian process upper confidence bound (UCB). 
The most optimal hyperparameter set would achieve the maximal EI~\cite{hutter2010sequential}. 
Bayesian optimisation demonstrates powerful performance on tuning convolutional~\cite{NIPS2012_4522} and fully-connected~\cite{mendoza2016towards} neural networks. 
However, the relationship between the covariance function and its associated hyperparameters is hard to be determined in practical problems~\cite{NIPS2012_4522}. 
Besides, these methods are based on prior over functions like Gaussian processes.
This requires an increasing number of data points to find better solutions in higher dimensional spaces~\cite{falkner2018bohb}.

Hyperparameter optimisation is also addressed within the scope of reinforcement learning, specifically for architectural network design~\cite{jomaa2019hyp}. In reinforcement learning, an agent is trained to learn a policy to take a better action by giving a reward for its action according to the current state. 
Hence, in hyperparameter tuning problem, the agent would be given a reward if the whole progress is shown to be maximising the objective values~\cite{Dong_2018_CVPR}. 
Recent studies have applied RL into tuning the hyperparameters for neural network~\cite{Dong_2018_CVPR,jomaa2019hyp} and have proven that reinforcement learning is suitable and outperforming than Bayesian optimisation for tuning dynamical hyperparameters~\cite{Dong_2018_CVPR}. 
Nevertheless, deep reinforcement learning systems are known to be very noisy~\cite{AAAI1816669}.
Various factors like the scale of reward, randomness in seed solution and environment dynamics could be affecting the sensitivity of the algorithm and result.

Apart from the above techniques, evolutionary algorithm is also recognised as a promising optimisation approach to have a comparable performance and shorten the execution time.  
Instead of searching the nearby optimal solution like Bayesian Optimisation, evolutionary algorithm is searching in a global space to pursue a better solution. 
After initialising the first population randomly and evaluating each individual via fitness function, the optimisation process evolves towards better regions of the search space through selection, crossover and mutation in each population. 
The solutions with better objective values are retained as much as possible to the next generation, and the evolution of the next population can guarantee exploration of high-potential new solutions. 
Some open source frameworks like DEAP~\cite{JMLR:v13:fortin12a}, provide practical tools for rapid prototyping of custom evolutionary algorithms, and are widely used for hyperparmater-tuning purposes/projects.
Recent experimental study shows that evolutionary algorithm has a significant improvement over the state-of-the-art~\cite{8297018,4378745,10.1016/j.neucom.2004.11.022,wu2007real} performances of machine learning.

Concurrent with our work, even though in the recent studies, the evolutionary algorithm has been shown to be compared favourably to the state-of-art hyperparameter tuning benchmark, the whole optimisation process would still take a long time, especially when the fitness function is expensive to execute.
Hence, in this paper, we propose a novel approach on the basis of the existing evolutionary optimisation algorithm, which can further shorten hyperparameter tuning execution time and speed up to achieve the convergence. 
A similar work was proposed by Smith et al.~\cite{6595766} by assembling Recurrent Neural Network into Surrogate-Assisted evolutionary optimisation to predict the convergence, such that the training of the surrogate model can be stopped early to reduce the total times of evaluating the expensive fitness function. Instead of predicting the convergence, \emph{IEO} predicts the occurrences of under-performing hyperparameters and skips unnecessary evaluation during the optimisation process so as to get benefit from saving evaluation time.

\section{The Solution Approach}
The proposed \emph{IEO} applies machine learning techniques to further shorten the optimisation process based on traditional evolutionary algorithm. 

When the traditional evolutionary algorithm is used to optimise hyperparameters, the initial populations, which do not necessarily include the optimal hyperparmater values, will be generated randomly. 
During the traditional optimisation process, in order to find the optimal solution, only the fitter solution will be passed to the next generation.
After the crossover and mutation process on the current generation, the next set of hyperparameters will be generated and fed into fitness function.
This process is continuously repeated until the optimal hyperparameters are found, that is, the objective values from fitness function are maximised/minimised.

Distinguished from traditional evolutionary algorithm, \emph{IEO} does not evaluate all individuals during the optimisation process.
Instead, in the optimisation process, \emph{IEO} uses a machine learning model to predict the selection of solutions between some generations during the process, that is, to determine whether the newly generated solutions are ``worthy'' to be evaluated by the fitness function before they are fed into it. 
If a solution is predicted as not having better objective value(s) than its parents, it would be assigned a \textit{null} value without actual evaluation and will be eliminated before the next generation. 
Thereby the total number of evaluations could be remarkably reduced and the optimisation duration could be shortened.
Since \emph{IEO} only uses machine learning technique to predict the performance of each individual, it does not affect the internal logic of the entire optimisation algorithm.

As for the criteria for determining whether the solution is good enough to evaluate, we compare the solution with its parent solutions. 
According to the fundamental idea of evolutionary algorithm, the offspring with the better performance have higher probability to survive and be selected into the next generation after modifying their genomes.
In \emph{IEO} estimator, each estimated solution is compared with its parent solutions to see if it outperforms or dominates its parents. 
Only the solution estimated as dominating its parents will be evaluated. 
Otherwise, for estimated dominated solutions, they will be regarded as less fit offspring and get \textit{null} value directly, because theoretically, they will not likely be used to form a new generation in the optimisation process.

\begin{figure}[tb]
\centering
\graphicspath{ {images/} }
\includegraphics[scale=0.36]{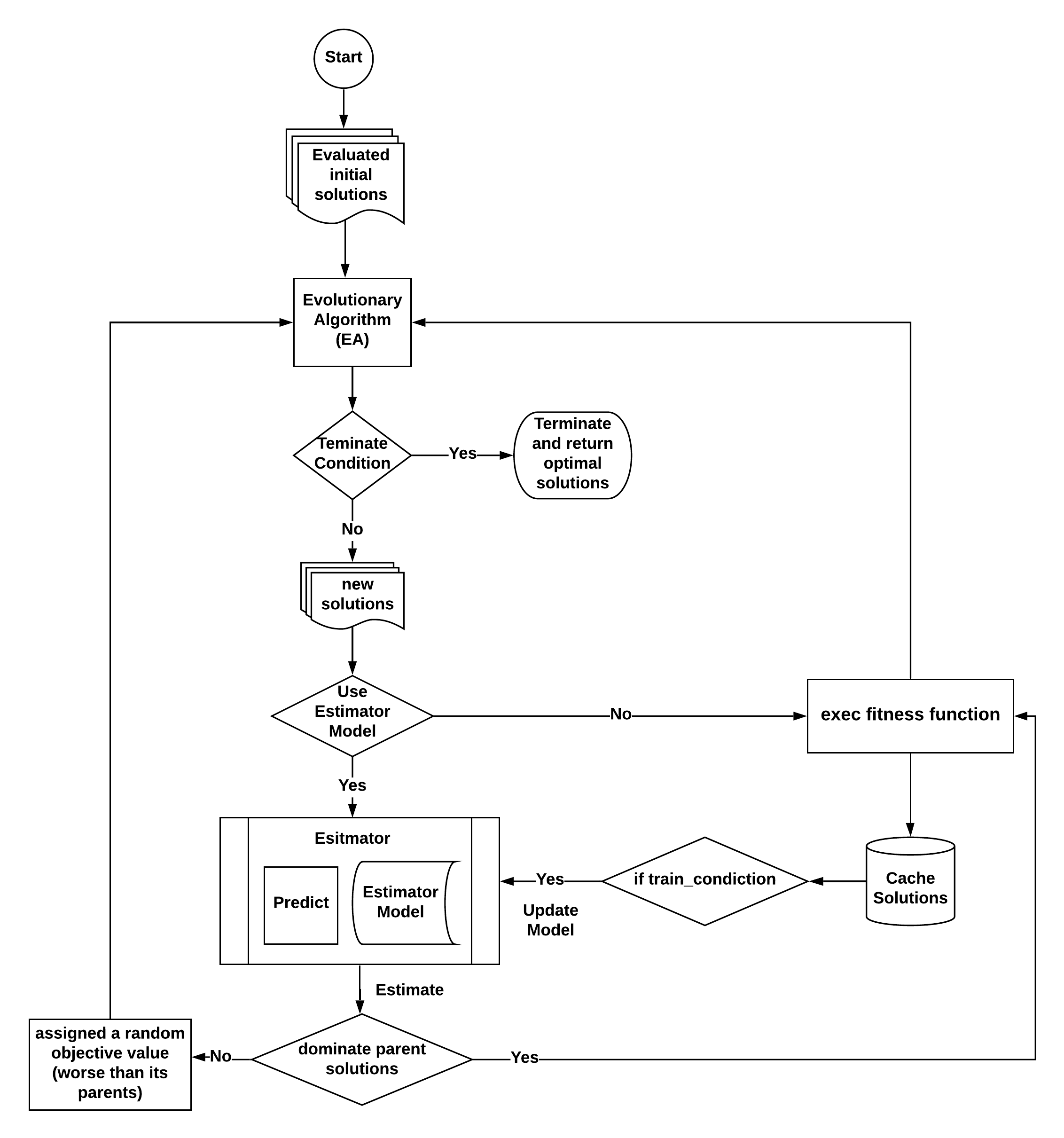}
\caption{IEO Architecture}
\end{figure}

\begin{algorithm}[tb]
\footnotesize
\caption{IEO main logic}
\label{nualgorithm}
\begin{algorithmic}
\REQUIRE solution, estimator\_model
\IF{estimate\_condition}
    \STATE est\_value = estimate(solution, estimator\_model)
    \IF{evaluate\_condition}
        \STATE objective\_values = evaluate(solution)
        \STATE historical\_solutions.append(solution)
    \ELSE
        \STATE objective\_values = null
\ENDIF
\ELSE
    \STATE objective\_values = evaluate(solution)
    \STATE historical\_solutions.append(solution)
    \IF{train\_condition}
        \STATE \# Update the model
        \STATE estimator\_model = train\_model(historical\_solutions)
    \ENDIF
\ENDIF
\end{algorithmic}
\end{algorithm}

\subsection{IEO estimator training details}
The core of \emph{IEO} is a binary classifier model. 
The primary idea of integrating a binary classifier is that after executing the fitness function for comparison and selection between generations a certain number of times, all the  historical solutions are used to generate a training dataset for training the base model of the estimator, to predict the likely performance of some subsequently generated solutions.

In terms of the training dataset format of the estimator, after a certain number of iterations, all the historical solution will be concatenated pairwise as training data.
To be more specific, we form [solution\_1, solution\_2, label] as a data point of training data, wherein the “solution\_1/2” represent the hyperparameters of the solution, and ``label'' indicates a comparison result between the two solutions. 
Based on the comparison outcome with two solutions, the label is set to 1 if solution\_1 dominates solution\_2, otherwise it is set to 0. 

In consideration of the data quality, at least the first 15\% of the historical solutions will be fed into fitness function rather than the estimator, to obtain their objective values for comparison and selection and to form the training datasets for training the predictor model of the estimator. 
Furthermore, in order to minimise the impact of overall performance, the predictor model of the estimator will be retrained every few generations during the entire optimisation process to ensure a relatively stable prediction performance of the model. 

\begin{algorithm}[tbp]
\footnotesize
\caption{Build/Update Estimator Model: \textbf{train\_model()}}
\label{nualgorithm}
\begin{algorithmic}
\REQUIRE all\_executed\_solutions
\FOR {solution s \textbf{in}\ all\_executed\_solutions}
\STATE label\_1 = s $\succ$ s.parent\_1
\STATE label\_2 = s $\succ$ s.parent\_2
\STATE train\_data.append([s\_features, s.parent\_1\_features, label\_1])
\STATE train\_data.append([s\_features, s.parent\_2\_features, label\_2])
\ENDFOR
\STATE Model = train(train\_data)
\RETURN Model
\end{algorithmic}
\end{algorithm}

\subsection{IEO estimator predict}

The estimator is triggered when a predictive model is built based on a certain number of evaluation results.
It predicts whether the newly generated solutions are better than their parents, hence whether worthy of being fed into evaluation process or not.
When an offspring solution is predicted as worse than its parent, the execution of the fitness function can be omitted. 
And this solution will be marked as the worst individual directly, for instance, assigning its objective value as \emph{null}, without being evaluated by fitness function or being added into the new generation.

\begin{algorithm}[tbp]
\footnotesize
\caption{Estimate solution: \textbf{evaluate()}}
\label{nualgorithm}
\begin{algorithmic}
\REQUIRE solution s
\STATE predict\_1\_feature = s\_features, s.parent\_1\_features
\STATE predict\_2\_feature = s\_features, s.parent\_2\_features
\STATE prediction\_1 = model(predict\_1\_feature)
\STATE prediction\_2 = model(predict\_2\_feature)
\IF{prediction\_1 == True \textbf{or} prediction\_2 == True}
    \RETURN True
\ELSE
    \RETURN False
\ENDIF
\end{algorithmic}
\end{algorithm}

\section{Empirical Study}
An experimental study is conducted on ten real-world classification datasets. 
Two approaches are compared: \emph{IEO} and a single-objective evolutionary algorithm (genetic algorithm). 
Henceforth, we will refer to the second approach as ``\emph{EA}'', which is well described in~\cite{whitley1994genetic}, and we will not expand the details of it in the paper.

\subsection{Datasets}
we utilise ten popular dataset listed in table 1 from UCI and OpenML, covering data for both binary and multiple classification. Each type contains data of different complexity, such as different number of features or data volumes.  Furthermore, for each type of data, we choose data in different qualities, which means that after we applied a basic machine learning model, we pick the data with different levels of test accuracy.

\begin{table}[tbp]
\centering
  \caption{This table shows the property of the datasets used in our experimental study.
  The \#Attributes, \#Instances, and \#Classes present the number of attributes, instances, and classes in each dataset, respectively.
  }
  \label{tb:dataset}
  \small
  \resizebox{0.45\textwidth}{!}{
   \begin{tabular}{llrrrr}
    \hline
    Name             & Source & \multicolumn{1}{l}{\#Attributes}  & \multicolumn{1}{l}{\#Instances}  & \multicolumn{1}{l}{Accuracy} & \multicolumn{1}{l}{\#Classes}  \\ \hline
    blood            & UCI    & 4                                 & 749                              & 83\%                         & 2                              \\
    gametes\_1       & OpenML & 19                                & 1,600                            & 79\%                         & 2                              \\
    gametes\_2       & OpenML & 19                                & 1,600                            & 69\%                         & 2                              \\
    spam\_test       & UCI    & 57                                & 4,601                            & 95\%                         & 2                              \\
    credit\_card     & UCI    & 24                                & 30,000                           & 82\%                         & 2                              \\
    heart\_diease    & UCI    & 18                                & 1,599                            & 98\%                         & 5                              \\
    thyroid          & OpenML & 27                                & 2,800                            & 76\%                         & 5                              \\
    gesture          & OpenML & 32                                & 9,873                            & 56\%                         & 5                              \\
    theorem-proving  & OpenML & 51                                & 6,118                            & 63\%                         & 6                              \\
    wine-quality-red & OpenML & 12                                & 2,800                            & 81\%                         & 7                              \\ \hline
    \end{tabular}
}
\end{table}


\subsection{Experimental Setup}
Experiments were performed in the environment of Ubuntu 18.04 LTS and Python 3.7, and the experiments were run in three Microsoft Azure F32s-v2 virtual machines with Intel Xeon Platinum 8272CL (Cascade Lake) processors and Intel® Xeon Platinum 8168 (Skylake) processors, featuring $32$ cores and $64$\,GiB of RAM.

Single objective genetic algorithm has been used for optimisation process, wherein Simulated Binary Crossover (SBC) with $100\%$ probability to generate next set of hyperparameter is selected as crossover method and Polynomina Mutation with $1/number\_of\_parameters$ probability is used for mutating the gene which refers to a single hyperparameter,  and 2-way tournament selection helps to choose the best individual from each tournament and keep them for the next crossover.

\subsection{Research Questions}
To evaluate the \emph{IEO} framework, we carry out an experimental study to assess the effectiveness and efficiency of this approach.
In the experiment, we demonstrate why we should integrate machine learning based estimator with the evolutionary algorithm in the process of hyperparameter tuning, 
and how much improvement can be achieved by employing \emph{IEO} as a hyperparameter tuning tool, thus raising two main research questions:

\textit{\textbf{RQ1:}} Can machine learning models effectively estimate the fitness of a solution?


We investigate the performance of estimator on predicting the quality of the given hyperparameter setting.
We analyse the performance quality by looking into the accuracy of predicting if the performance of a hyperparameter setting is dominated by another hyperparameter setting.
This research question is a foundation for applying \emph{IEO} on hyperparameter tuning problem.

\textit{\textbf{RQ2:}} What is the effectiveness and efficiency of \emph{IEO} in the process of hyperparameter tuning?


This question can be expressed in a quantified manner as to how much performance improvement can be obtained by \emph{IEO} compared with the \emph{EA}.
This question is composed of two more detailed sub-questions (\textbf{\textit{RQ2.1}}, and \textbf{\textit{RQ2.2}}):

\textit{\textbf{RQ2.1: }} Can our approach improve the efficiency of the hyperparameter tuning progress? 
To understand the efficiency of the \emph{IEO}, we compare the convergence of \emph{IEO} with \emph{EA} in terms of two convergence metrics (execution time and iteration numbers). 

\textit{\textbf{RQ2.2: }} Does \emph{IEO} affect the performance quality of the final optimisation result?
We statistically measure the performance difference between the optimal solutions generated by \emph{IEO} and \emph{EA} to see whether \emph{IEO} can explore nearly good or even better hyperparameter settings.
    
    

\section{Experiment Results}
In this section, we present the results of the experimental study, and interpret the research questions sequentially and separately to explain why \emph{IEO} is better than traditional evolutionary algorithm for hyperparameter tuning. 

\subsection{RQ1: Can machine learning models effectively estimate the fitness of a solution?}
In order to answer this question, we apply different classification machine learning models on historical solutions to understand the ability of these machine learning models on learning the pattern between hyperparameters and model performance. 
We collect all generated solutions from optimisation process, and match them pairwise to form dataset as [solution\_1, solution\_2, label] wherein label represents the non-dominated sorting result of two solutions.
We use non-dominated sorting approach here because this enables the ability of applying the proposed approach on multi-objective based hyperparameter tuning.
Based on the mechanism of pairwise comparison, there are $k*(k-1)/2$ comparisons generated from $k$ historical solutions.
And the average ratio of label $1$ and label $0$ are $0.64$ and $0.36$, respectively.
The reason why the percentage of two labels are not equal is that there are some comparisons that are non-dominated.
After we generated the dataset and splitted the data into training set and testing set, we then apply various machine learning models to validate whether the models are able to estimate the performance of a machine learning model by giving an unseen hyperparameter setting. 
To understand the possible cost of introducing the proposed approach in hyperparameter tuning process, we also record the execution time of model training. 

\textbf{RQ1 can be answered by Table~\ref{tb:rq1}}, which illustrates the accuracy scores and execution times of different machine learning models.

\begin{table*}[tbp]
\centering
  \caption{This table compares the accuracy scores and execution times of different machine learning models applying on historical data comes out from evolutionary optimisation.
  }
  \label{tb:rq1}
  \small
  \resizebox{0.98\textwidth}{!}{
    \begin{tabular}{@{}lllllllllll@{}}
\toprule
                 &                     & Randomforest                & Xgboost                     & Extratree          & SVM               & Adaboost          & GaussianProcess    & Kneighbours       & GradientBoosting  & MLP               \\ \midrule
blood            & train acc/ test acc & 99.19\% / 83.96\%           & 97.29\%\textbf{/ 93.94\%}  & 100.00\% / 81.30\% & 85.70\% / 68.19\% & 78.10\% / 78.70\% & 89.54\% / 69.15\%  & 78.73\% / 70.87\% & 83.36\% / 82.60\% & 71.19\% / 69.69\% \\
                 & exec time (s)       & 0.67                        & 2.21                        & \textbf{0.56}      & 26.08             & 0.86              & 325.86             & 1.05              & 2.82              & 4.62              \\
credit card      & train acc/ test acc        & 99.66\% / 86.42\%           & 97.64\%\textbf{ / 92.71\%}  & 100.00\% / 83.22\% & 94.80\% / 70.07\% & 81.83\% / 81.36\% & 98.57\% / 71.86\%  & 84.85\% / 76.42\% & 85.18\% / 84.09\% & 76.66\% / 75.46\% \\
                 & exec time (s)       & 0.64                        & 2.23                        & \textbf{0.56}      & 22.93             & 1.04              & 251.05             & 1.03              & 3.93              & 3.94              \\
spam test        & train acc/ test acc           & 99.74\%\textbf{ / 100.00\%} & 100.00\% / 99.31\%          & 100.00\% / 94.68\% & 99.17\% / 71.43\% & 90.57\% / 90.58\% & 100.00\% / 82.37\% & 90.84\% / 84.94\% & 94.45\% / 94.00\% & 87.65\% / 85.70\% \\
                 & exec time (s)       & 0.66                        & 2.19                        & \textbf{0.46}      & 20.86             & 1.11              & 260.91             & 1.03              & 4.11              & 12.56             \\
Gametes 1        & train acc/ test acc           & 99.64\% / 92.71\%           & 99.98\% \textbf{ / 99.19\%}  & 100.00\% / 99.34\% & 98.13\% / 70.85\% & 84.79\% / 84.03\% & 99.94\% / 79.57\%  & 87.93\% / 82.46\% & 90.17\% / 89.55\% & 83.95\% / 82.28\% \\
                 & exec time (s)       & 0.67                        & 2.17                        & \textbf{0.56}      & 26.07             & 1.34              & 254.21             & 1.02              & 3.76              & 13.66             \\
Gametes 2        & train acc/ test acc          & 99.60\% / 95.73\%           & 100.00\% \textbf{/ 99.75\%} & 100.00\% / 94.23\% & 75.14\% / 75.59\% & 89.50\% / 90.04\% & 99.02\% / 73.71\%  & 87.81\% / 81.70\% & 92.90\% / 93.10\% & 82.30\% / 82.33\% \\
                 & exec time (s)       & 0.66                        & 2.07                        & \textbf{0.56}      & 21.69             & 0.99              & 254.43             & 1.11              & 3.65              & 7.95              \\
heart diease     & train acc/ test acc           & 99.63\% / 90.92\%           & 98.56\% \textbf{/ 95.30\%}  & 100.00\% / 89.19\% & 93.28\% / 72.46\% & 84.28\% / 83.62\% & 95.74\% / 74.83\%  & 84.49\% / 77.20\% & 89.01\% / 88.64\% & 83.48\% / 80.78\% \\
                 & exec time (s)       & 0.67                        & 2.31                        & \textbf{0.56}      & 23.47             & 0.99              & 328.43             & 1.12              & 3.5               & 5.93              \\
gesture          & train acc/ test acc           & 99.84\% / 96.38\%           & 99.59\% \textbf{ / 98.08\%}  & 100.00\% / 95.73\% & 92.85\% / 87.65\% & 94.36\% / 94.88\% & 95.74\% / 91.59\%  & 94.98\% / 93.15\% & 95.69\% / 95.44\% & 88.37\% / 88.90\% \\
                 & exec time (s)       & 0.57                        & 2.17                        & \textbf{0.56}      & 17.9              & 0.78              & 329.03             & 1.43              & 2.24              & 0.45              \\
thyroid          & train acc/ test acc           & 99.47\% /85.75\%            & 96.10\% \textbf{/ 91.61\%}  & 100.00\% / 82.82\% & 86.40\% / 76.60\% & 81.83\% / 82.19\% & 91.39\% / 76.67\%  & 84.64\% / 77.96\% & 84.96\% / 85.08\% & 76.54\% / 77.29\% \\
                 & exec time (s)       & 0.67                        & 2.18                        & \textbf{0.56}      & 23.11             & 0.88              & 289.87             & 1.86              & 2.94              & 7.13              \\
wine             & train acc/ test acc           & 99.60\% / 90.16\%           & 98.05\% \textbf{/ 94.12\%}  & 100.00\% / 88.10\% & 90.96\% / 79.35\% & 87.51\% / 86.31\% & 94.73\% / 81.79\%  & 88.68\% / 83.83\% & 88.99\% / 88.21\% & 80.82\% 78.99\%   \\
                 & exec time (s)       & 0.67                        & 2.23                        & \textbf{0.56}      & 18.92             & 1.06              & 293.88             & 1.51              & 3.63              & 6.6               \\
theorem proving  & train acc/ test acc           & 99.38\% / 87.05\%           & 96.45\% \textbf{/ 92.51\%}  & 100.00\% / 84.99\% & 89.34\% / 76.10\% & 84.88\% / 84.56\% & 92.00\% / 81.45\%  & 88.15\% / 81.86\% & 86.58\% / 86.53\% & 72.16\% / 71.12\% \\
                 & exec time (s)       & 0.67                        & 2.15                        & \textbf{0.56}      & 17.89             & 0.82              & 294.05             & 1.16              & 2.63              & 6.41              \\
overall rankiing & train acc/ test acc           & 2.5/2                         & 2.6/\textbf{1.2}                         & \textbf{1}/3.3         & 5.3/8.8               & 7.8/5               & 3.5/7.7                & 6.9/6.3               & 6/3.5                 & 9/7.2                 \\
                 & exec time (s)       & 2.1                         & 5.1                         & \textbf{1.1}       & 8                 & 3.4               & 9                  & 3.8               & 6.1               & 6.4               \\ \bottomrule
\end{tabular}
}
\end{table*}

We apply 9 different most well known classification machine learning models (such as XGBoost Classifier, SVM, and MLP classifier) on selected dataset, to have the best possible variety. 
As is shown in Table~\ref{tb:rq1}, among all the dataset, the test accuracy scores of selected machine learning models are between 68.19\% and 100.00\%. 
Besides, most of the models can finish the train and predict process in less than 3 seconds, except for SVM and GaussianProcess.

Comparing all the models, we can see that overall XGBoost Classifier has the best predictive performance on all the dataset with a stable prediction accuracy rate of at least 91.61\%, and up to 99.75\%. 
Besides, under the premise of ensuring a certain accuracy rate, XGBoost Classifier has a relative short execution time. 
On average, it takes up to 2.5s to train the model and around 0.01s to predict the result. 
During the optimisation process, depends on the expense of fitness function and the size of dataset, each evaluation process usually takes a few seconds or much longer. Some individuals which are predicted as having poor performance would not be fed into the fitness function, its corresponding evaluation process can be skipped, and considerable processing time can still be saved.
Therefore, although XGBoost Classifier is not one of the most time-saving model, considering the better predictive performance, we apply it as the base model of \emph{IEO} optimisation process.
In other words, compared with the time saved in the entire optimisation process, the time cost of training process for XGBoost Classifier is negligible and acceptable.

Therefore, from the experiment, we can conclude that machine learning can effectively estimate the performance of the machine learning model in general and XGBoost Classifier could be selected as the base model in our \emph{IEO} estimator to reduce the deviation of the experimental results due to the unstable performance of the estimator model.

\subsection{RQ2: What is the effectiveness and efficiency of \emph{IEO} in the process of hyperparameter tuning?}

To answer this question, we ran a series of controlled experiments to investigate the effectiveness and efficiency of \emph{IEO}. 
Three popular classification machine learning models (Xgboost Classifier, Randomforest Classifier and Label Spreading Classifier) are used as subjects in the experimental study.
The experimental study is undertaken under the same termination criteria (same iterations number) and the same dateset.
The only difference is the algorithms used for hyperparameter optimisation.
In order to reduce bias and avoid the contingency of experiment results, the experiments across all the selected dataset are repeated 30 times.
We normalise the experiment result by the maximal execution time in each comparison ($normalised\_val = val / max\_val$) to intuitively demonstrate the difference. 
Specifically, in each comparison, the maximum value among 30 results of \emph{EA} gets transformed into 1 and every other values of \emph{EA} and \emph{IEO} are transformed into a decimal between 0 and 1.
Additionally, we use the Wilcoxon Test and Student's t-test to observe the statistical power between the controlled experiments.

\textit{\textbf{RQ2.1}} Can our approach improve the efficiency of the hyperparameter tuning progress? 

In response to this question, we compared the overall time taken by the \emph{EA} and \emph{IEO} approach.

\begin{figure}[t!]
\begin{subfigure}{.5\textwidth}
  \centering
  \includegraphics[width=.95\linewidth]{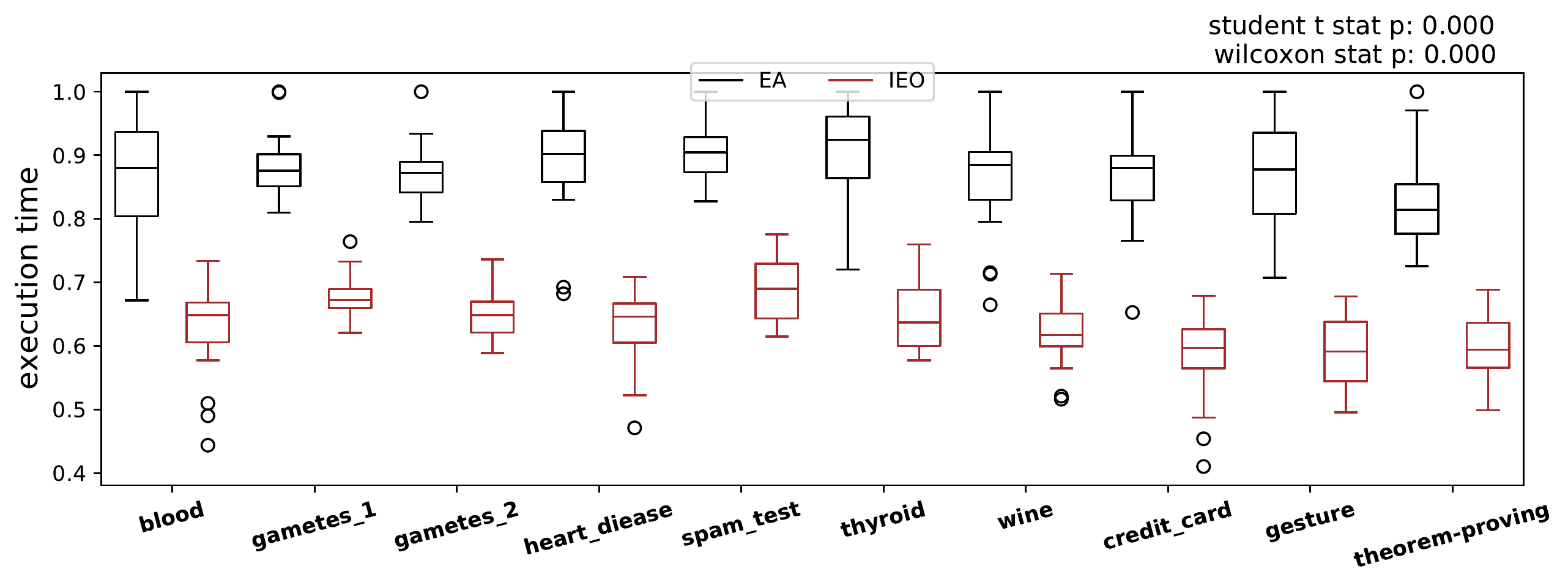}  
  \caption{Randomforest Classifier}
  \label{fig:sub-first}
\end{subfigure}
\begin{subfigure}{.5\textwidth}
  \centering
  \includegraphics[width=.95\linewidth]{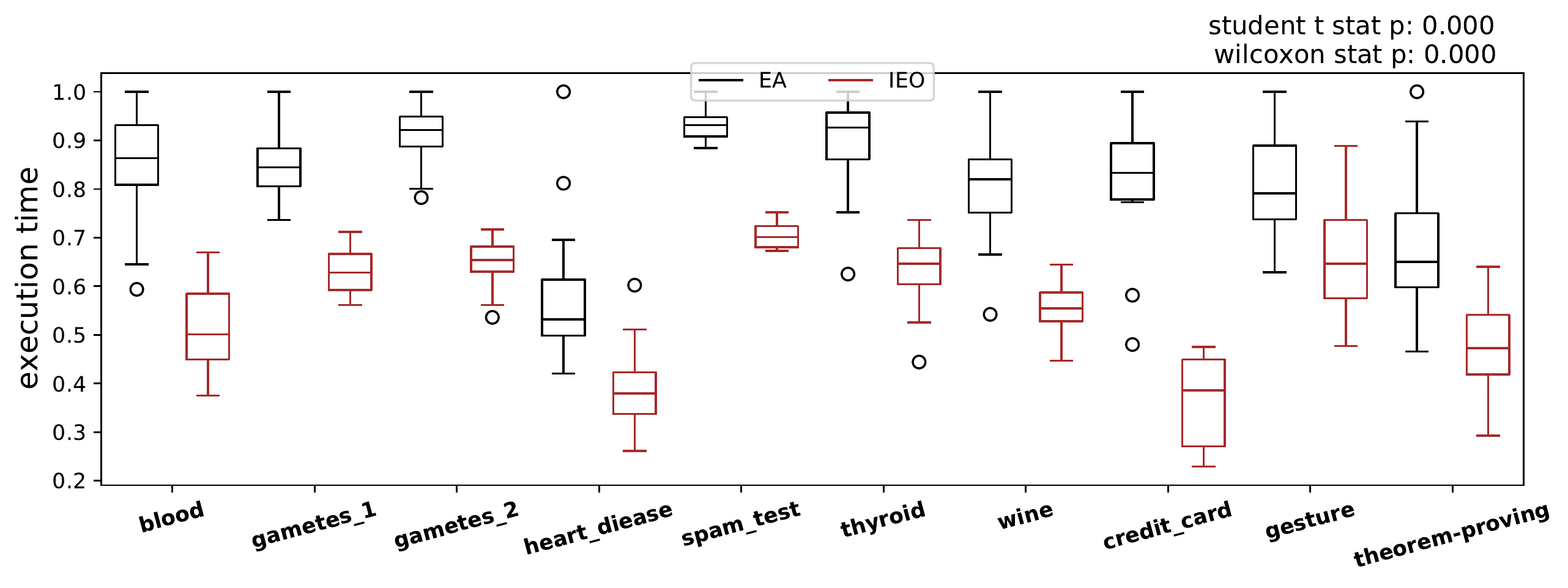}  
  \caption{Label Spreading Classifier}
  \label{fig:sub-second}
\end{subfigure}
\begin{subfigure}{.5\textwidth}
  \centering
  \includegraphics[width=.95\linewidth]{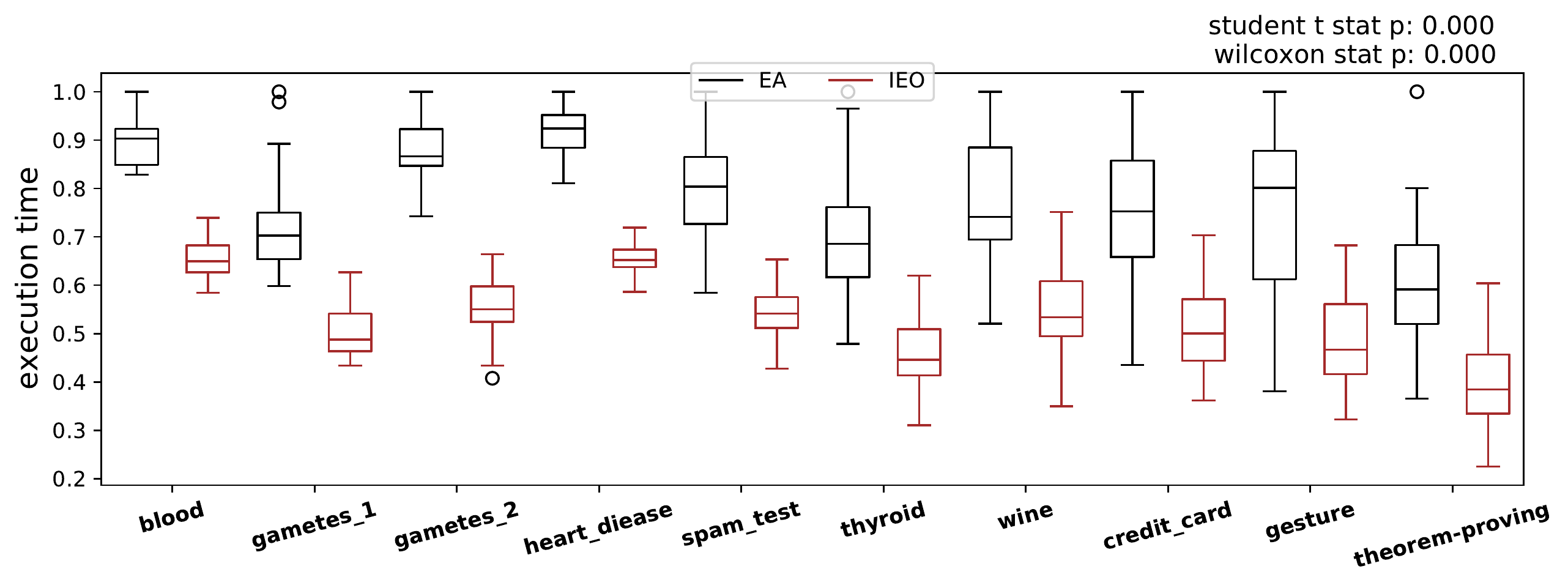}  
  \caption{Xgboost Classifier}
  \label{fig:sub-second}
\end{subfigure}
\caption{Optimisation process execution time. The lower, the better.}
\label{execution time}
\end{figure}

Figure~\ref{execution time} shows that our \emph{IEO} effectively reduces the optimisation time in all subjects.
Through statistical experiment results, with our \emph{IEO} optimisation mechanism, the entire optimisation process speeds up about 30.40\%, and up to 77.06\% improvement among three target machine learning models in the best scenarios. 
Besides, the average p-value is much less than 0.0001, which indicates strong evidence that our improvement is statistically significant.
This result also proves the hypothesis in RQ1 that comparing with the whole process of optimisation, the training time of base model would not make a big impact and could be neglected.
It should be noted that the current proportion of saved time could be affected by a few factors, such as the number of times the fitness function is actually executed before starting the estimator, the setting of how many individuals are estimated during the process and the times to retrain the \emph{IEO} model. Those influencing factors lead to a case by case result.

\begin{figure}[t!]
\begin{subfigure}{.5\textwidth}
  \centering
  \includegraphics[width=.95\linewidth]{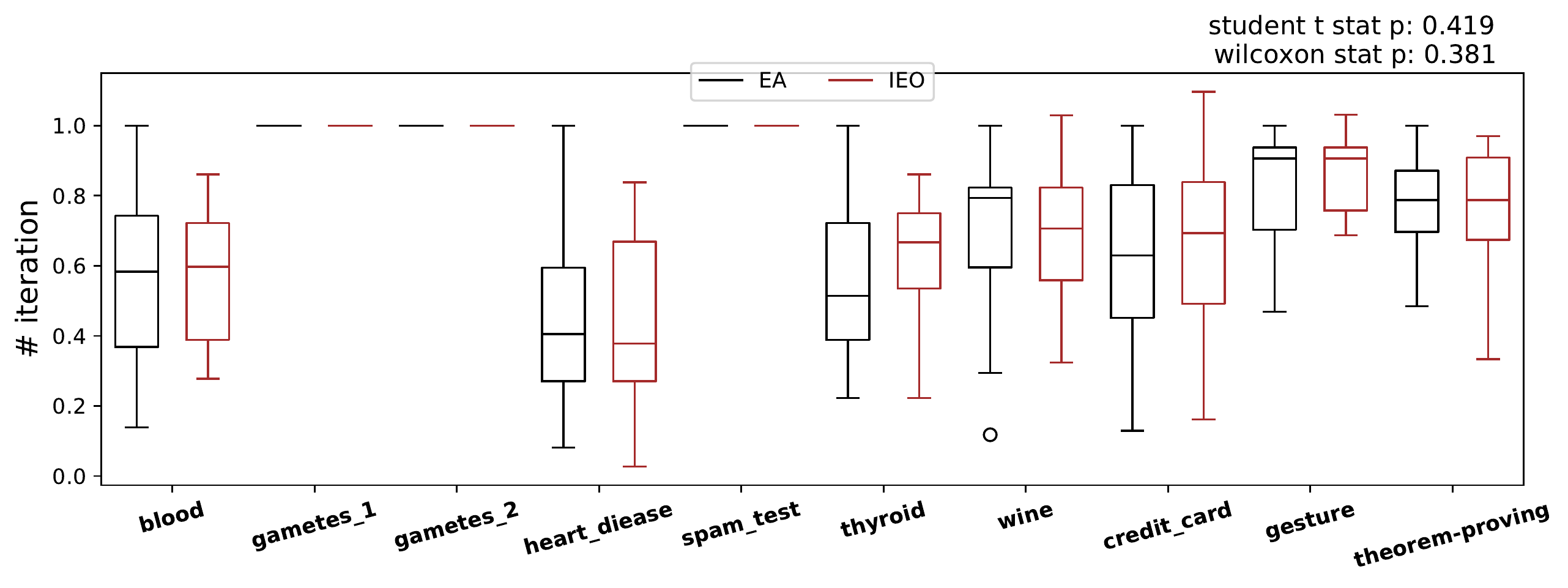}  
  \caption{Randomforest Classifier}
  \label{fig:sub-first}
\end{subfigure}
\begin{subfigure}{.5\textwidth}
  \centering
  \includegraphics[width=.95\linewidth]{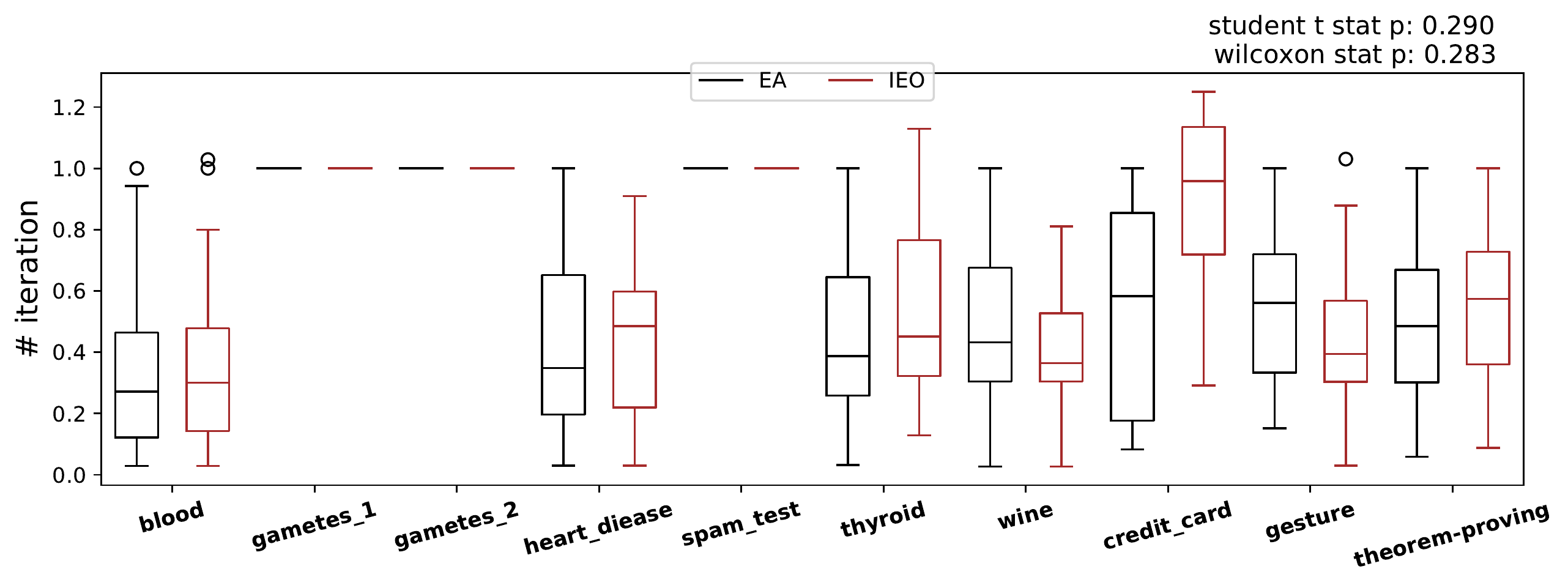}  
  \caption{Label Spreading Classifier}
  \label{fig:sub-second}
\end{subfigure}
\begin{subfigure}{.5\textwidth}
  \centering
  \includegraphics[width=.95\linewidth]{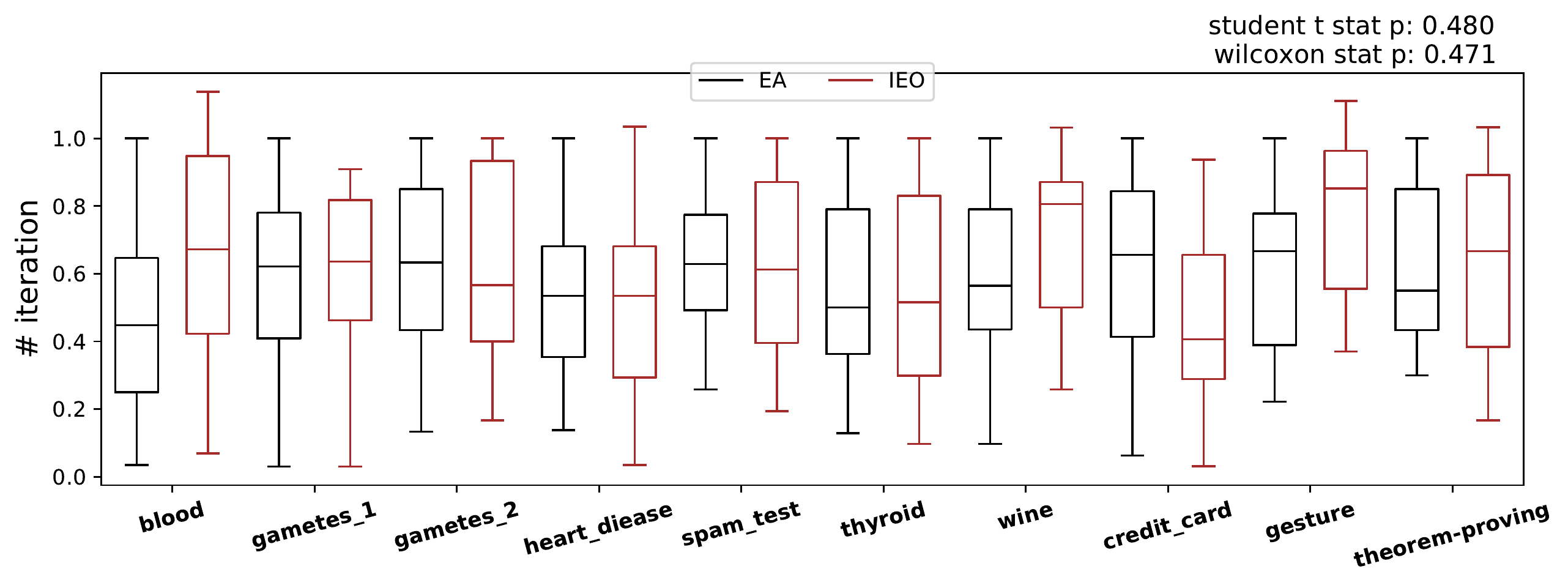}  
  \caption{Xgboost Classifier}
  \label{fig:sub-second}
\end{subfigure}
\caption{Convergence speed in terms of the number of iteration. The lower, the better.}
\label{convergence}
\end{figure}

In terms of the convergence time, we calculated the number of iterations when the objective function reaches its maximum (or minimum) value. As shown in Figure~\ref{convergence}, in general, while improving the optimisation time, the convergence rate of \emph{IEO} is comparable with \emph{EA}. In addition, among three models, the Wilcoxon statistical p-values of the difference are 0.381, 0.283 and 0.471 respectively, which indicate there is no statistical significant between the result. 

Therefore, while keeping the convergence speed, the entire optimisation process can save at least a few minutes and up to hundreds of minutes, depending on the difficulty of the data set and the complexity of the model to be tuned, which proves that \emph{IEO} reduces the time cost of optimisation process significantly (\textbf{this answers \emph{RQ2.1}}).

\textit{\textbf{RQ2.2}} Does \emph{IEO} affect the performance quality of the final optimisation result?

\begin{figure}[t!]
\begin{subfigure}{.5\textwidth}
  \centering
  \includegraphics[width=.95\linewidth]{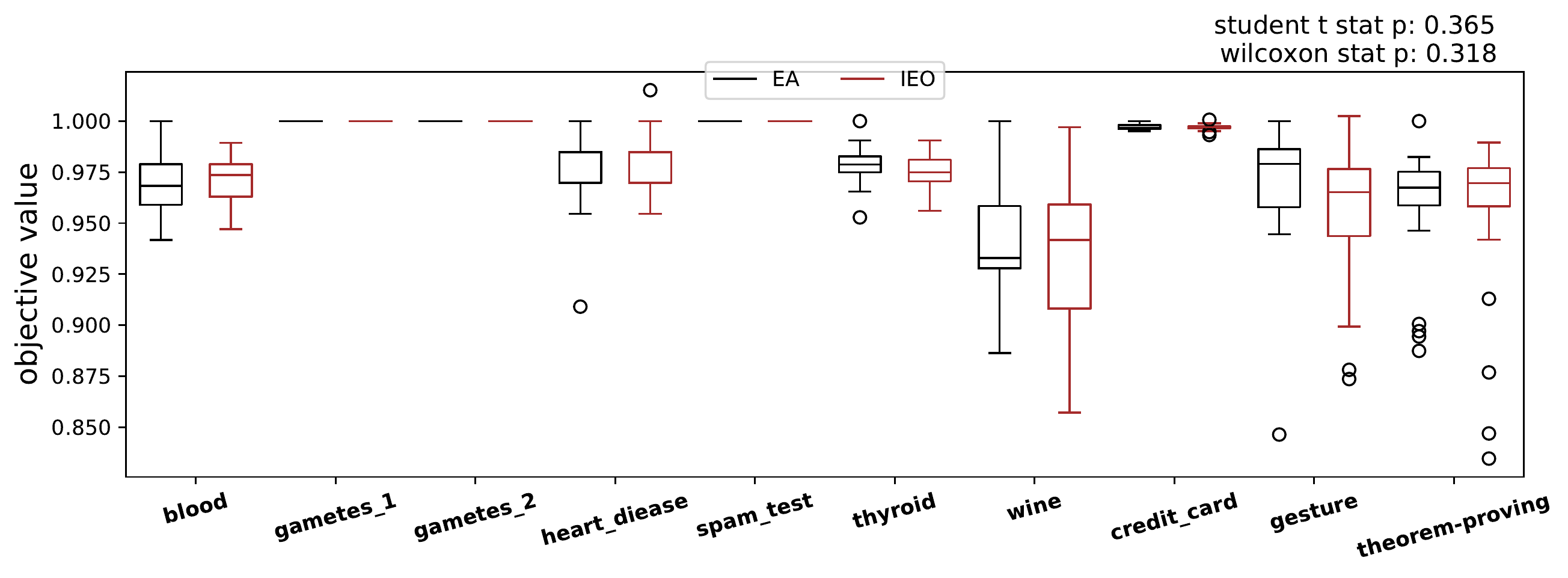}  
  \caption{Randomforest Classifier}
  \label{fig:sub-first}
\end{subfigure}
\begin{subfigure}{.5\textwidth}
  \centering
  \includegraphics[width=.95\linewidth]{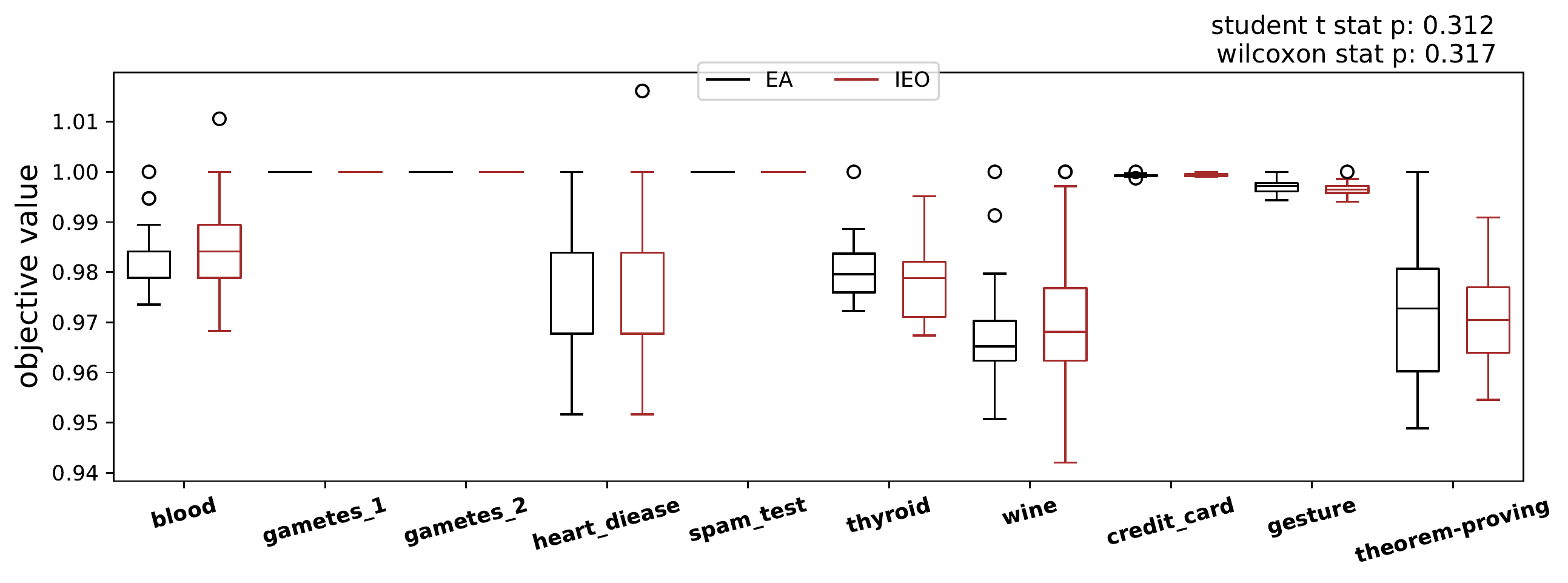}  
  \caption{Label Spreading Classifier}
  \label{fig:sub-first}
\end{subfigure}
\begin{subfigure}{.5\textwidth}
  \centering
  \includegraphics[width=.95\linewidth]{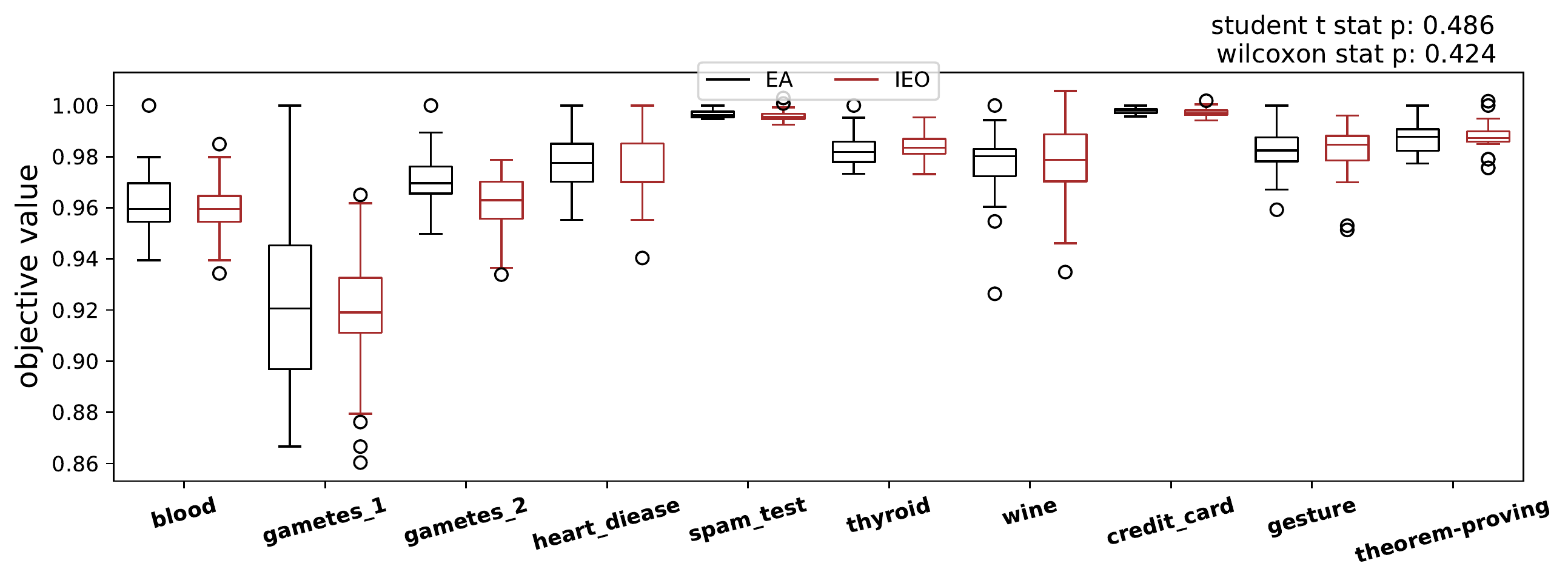}  
  \caption{Xgboost Classifier}
  \label{fig:sub-first}
\end{subfigure}
\caption{Optimality of the optimal solutions. The higher, the better.}
\label{max value}
\end{figure}

To answer this question, we present the comparison between objective values of optimal solutions respectively obtained by \emph{EA} and \emph{IEO} under the same experimental configuration. 
For this, we use \emph{EA} and \emph{IEO} to run each hyperparameter optimisation experiment for each dataset on the same model for 30 times, and statistically compare the optimal solutions obtained by the two algorithms to verify whether our algorithm \emph{IEO} has a comparable performance as the well known \emph{EA}.
From Figure~\ref{max value} and the statistical test result, it is clear that the difference between the benchmark and \emph{IEO} is negligible.

From all the experiments' results, the Wilconxon statistical p-value of the difference of three models are 0.365, 0.312 and 0.486, which is much higher than 0.05 and referring strong evidence that there is no statistically significant between \emph{EA} and \emph{IEO}'s performance.
Though \emph{IEO} skips the evaluation part for some under-performed solutions during the optimisation process, it still guarantees the overall performance quality comparing with the \emph{EA}. 
Accordingly, it illustrates that \emph{IEO} can effectively predict the performance of the solution by comparing its parent solution with itself.

From the above experiment result, it shows that our approach could improve the efficiency of evolutionary algorithm optimisation without affecting the original optimisation's performance, which \textbf{answers \emph{RQ2.2}}.

\section{Threats to validity}
One threat may come from the experiment environment.
When comparing the efficiency of \emph{IEO} and \emph{EA}, we compared the execution time of the entire optimisation process. 
Different device configuration and memory usage will affect the experiment result. 
We mitigate this threat by running the experiments on independent azure machines, excluding all the other system application to make sure the consistency of the experimental environment. 

Another threat may come from the characteristics of algorithm we use.
In our experimental study, genetic algorithm was selected as the control group. 
We select it as a base line since it is one of the most popular and advanced evolutionary algorithm. 
The stochastic characteristics of optimisation algorithms may lead to the possibility of diversification of results.

In addition, biases in datasets, like selection biases or recall bias can produce misleading result. 
The missing values, imbalanced classification or other biases will interfere with the accurate prediction of the model and increase the difficulty of hyperparameter optimisation, thereby standing in the way of getting accurate result of controlled experiments. 
We mitigated it by increasing the diversity of dataset, selecting various dataset from multiple perspectives.
Also, in order to deal with the missing values or lower representative information of the dataset, we preprocessed the data before running the experiments, such as cleaning up, normalising and feature selection to reduce the interference caused by bias as much as possible.

Other than that, the stochastic characteristics of evaluating models is also a threat. 
The uncertainty can come from two aspects: the randomness of model's performance and the model selected may rarely be able to capture all of the aspects of the domain. 
We attempt to mitigate this threat by running 30 times of each experiment and choosing 3 different model to increase the diversity of the aspects of the domain.

\vspace*{-1em}
\section{Conclusion \& Future work}
In this paper, we proposed an intelligent evolutionary optimisation algorithm which applies machine learning technique into traditional evolutionary algorithm to accelerate the hyperparameter optimisation process of classification machine learning models. 
Specifically, the proposed approach \emph{IEO} improves the efficiency of the optimisation process by skipping unnecessary model training via learning the pattern of under-performed hyperparameters during the optimisation process.
In a series of controlled experiments involving 10 various open datasets and 3 popular machine learning models, we empirically proved that our \emph{IEO} is capable to reduce the hyperparameter tuning optimisation time by 30.40\% on average and 77.06\% in the best scenarios. 
Besides, after comparing the performance of traditional \emph{EA} and our \emph{IEO} approach, we found out that \emph{IEO} is not affecting the optimal value of targeted models while speeding up the hyperparameter tuning optimisation process.
In conclusion, while keeping performance at a comparable level, \emph{IEO} significantly improves the efficiency of hyperparameter optimisation effectively.

In the future, a research of neural network hyperparameter tuning is required to gradually validate the generalisation of \emph{IEO}. 
Besides, another direction of future work is investigating how to improve both effectiveness and efficiency of the hyperparameter tuning.




\pagebreak
\bibliography{sample-sigconf} 

\end{document}